%% file: main.tex
\documentclass[]{bytedance_seed}

\usepackage[toc,page,header]{appendix}

\usepackage{minitoc}
\usepackage{CJKutf8}
\usepackage{wrapfig}
\usepackage{lipsum}
\usepackage{wrapfig}
\usepackage{subcaption}

\input{math_commands.tex}

\usepackage{hyperref}
\usepackage{url}
\usepackage{multicol}
\usepackage{aliascnt}
\usepackage{adjustbox}
\usepackage{tcolorbox}
\usepackage{booktabs}
\usepackage{multirow} 
\usepackage{todonotes}  
\usepackage{enumitem}
\usepackage{float}
\usepackage{multirow}
\usepackage[table]{xcolor} 
\usepackage{booktabs}

\usepackage{graphicx} 
\usepackage{subcaption}

\title{\centering Beyond Correctness: Evaluating Subjective Writing Preferences Across Cultures}

\affiliation{ByteDance Seed, M-A-P}

\contribution{Full author list in Contributions}

\abstract{
Current preference learning methods achieve high accuracy on standard benchmarks but 
exhibit significant performance degradation when objective quality signals are removed. 
We introduce \textbf{WritingPreferenceBench}, a dataset of 1,800 human-annotated 
preference pairs (1,200 English, 600 Chinese) across 8 creative writing genres, where 
responses are matched for objective correctness, factual accuracy, and length. On this 
benchmark, sequence-based reward models—the standard architecture for RLHF—achieve only 
52.7\% mean accuracy, while zero-shot language model judges perform at 53.9\%. 
In contrast, generative reward models that produce explicit reasoning chains achieve 81.8\% accuracy. We observe high within-model variance across genres: individual models range from 18.2\% to 81.8\% accuracy across different writing categories, with standard deviations averaging 10.1\%. 
This variance persists regardless of model scale, with 27B parameter models showing no consistent improvement over 8B variants. Our results suggest that current RLHF methods primarily learn to detect objective errors rather than capture subjective quality preferences (e.g., creativity, stylistic flair, and emotional resonance), and that successful preference modeling may require intermediate reasoning representations rather than direct classification.
}

\date{\today}

\correspondence{Ge Zhang at \email{zhangge.eli@bytedance.com}}
\checkdata[Project Page]{\url{https://WritingPreferenceBench.github.io/}}

\begin{document}

\maketitle

\section{Introduction}

Reinforcement learning from human feedback (RLHF) has become the dominant paradigm for aligning language models with human values~\citep{christiano2017deep,ouyang2022training,bai2022constitutional}. Reward models trained via RLHF achieve 95\% accuracy on RewardBench's objective tasks~\citep{lambert2024rewardbench}, which emphasize safety violations, factual errors, and instruction-following. However, our benchmark reveals a critical limitation: when we systematically remove objective quality signals (grammatical errors, factual mistakes, length differences), sequence-based reward models—the dominant architecture in production RLHF systems—collapse to 52.7\% accuracy on writing preference tasks, barely above random chance. This 42-percentage-point degradation indicates that current preference learning primarily optimizes for error detection rather than recognition of subjective creative quality—a fundamental misalignment between training objectives and the aesthetic judgment required for creative tasks.

However, writing tasks constitute over 40\% of language model interactions~\citep{openai2025usage,anthropic2025economic}, spanning creative fiction, persuasive essays, and personal expression where subjective quality matters more than objective correctness. Yet our evaluation infrastructure remains anchored in verifiable metrics. RewardBench~\citep{lambert2024rewardbench} conflates safety with preference; WritingBench mixes creative with functional tasks~\citep{wu2025writingbench}; LitBench uses Reddit upvotes as quality proxies~\citep{fein2025litbenchbenchmarkdatasetreliable}. Moreover, existing benchmarks predominantly focus on English, leaving cross-lingual preference evaluation, particularly for languages with distinct rhetorical traditions like Chinese, largely unexplored. Recent theoretical work warns of ``reward hacking'' where models exploit spurious correlations rather than learning genuine preferences~\citep{pan2022effects}. 

\begin{figure}[t]
\centering
    \includegraphics[width=0.9\textwidth]{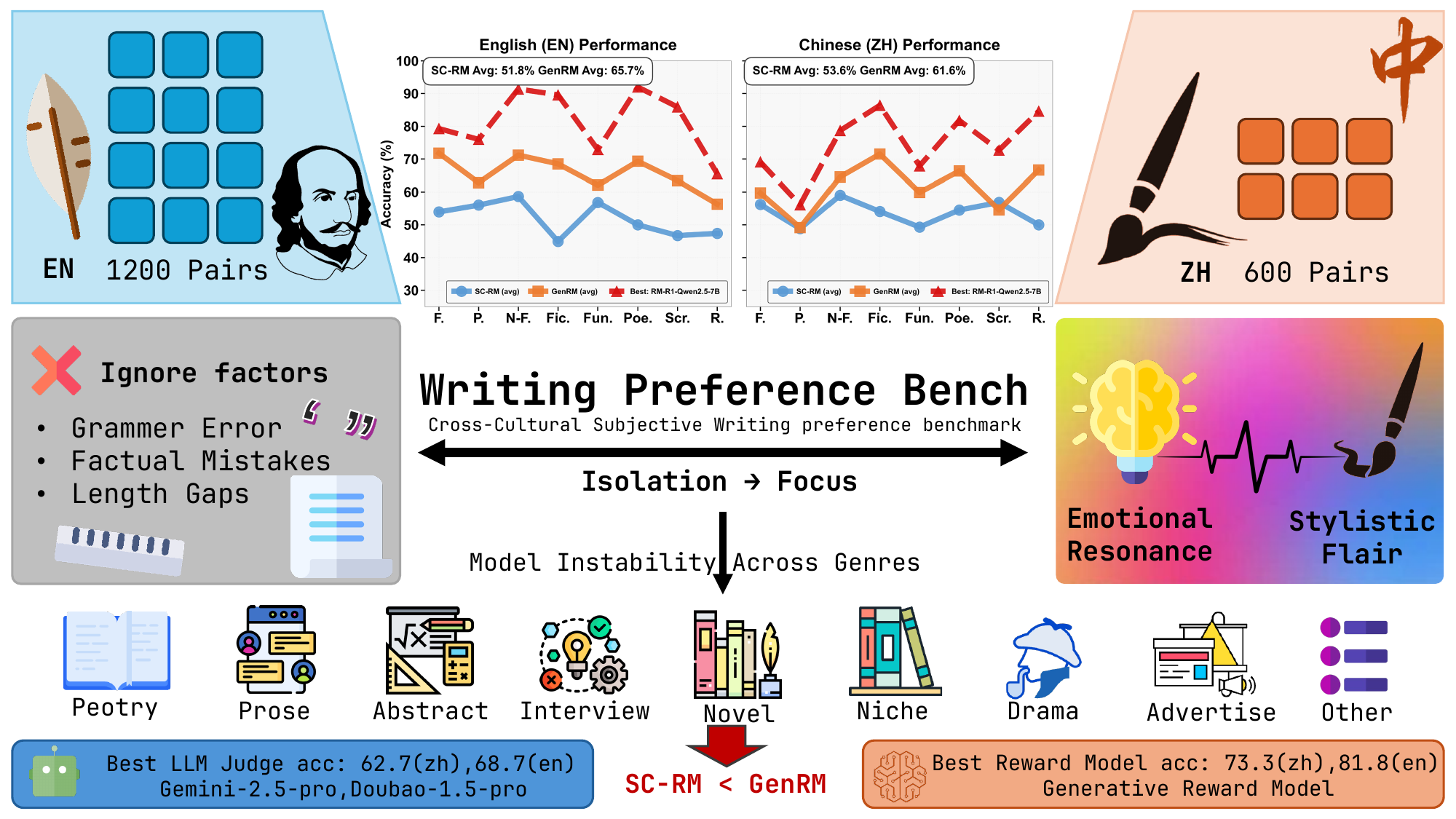}
    \caption{\textbf{WritingPreferenceBench} isolates subjective writing quality by neutralizing objective confounds (grammar, factuality, length). Across 1,800 human-validated preference pairs, standard sequence classifiers (SC-RM) perform near-randomly while generative reward models (GenRM) achieve 30\% higher accuracy—but both architectures exhibit catastrophic instability across genres, exposing the brittleness of current preference learning.}
\label{fig:teaser}
\end{figure}

We introduce \textbf{WritingPreferenceBench}, a cross-lingual dataset of 1,800 human-annotated preference pairs (1,200 English, 600 Chinese) across 8 creative writing genres
where both responses are grammatically correct, factually accurate, and length-matched. We focus on three dimensions of creative quality: \emph{creativity} (original ideas and novel perspectives), \emph{stylistic sophistication} (narrative techniques and linguistic elegance), and \emph{emotional resonance} (capacity to evoke authentic responses). By neutralizing objective confounds, our benchmark tests whether models can recognize these aesthetic qualities that distinguish compelling from merely competent writing.

Our evaluation of 21 models, comprising 7 reward models and 14 language models as judges spanning open-source and proprietary families, reveals fundamental limitations in current preference learning approaches. Sequence-based reward models, the dominant architecture in production RLHF systems~\citep{rafailov2023direct}, achieve only 52.7\% mean accuracy across both languages, while zero-shot language model judges~\citep{zheng2023judging} reach 53.9\%. Both results are statistically indistinguishable from random chance. In stark contrast, generative reward models that produce explicit reasoning chains~\citep{chen2025rm} achieve 81.8\% accuracy. This 29-percentage-point gap indicates that subjective preference modeling requires structured intermediate reasoning rather than direct pattern matching. Moreover, all architectures exhibit severe genre instability (mean standard deviation 10.9\%), with individual models ranging from 18.2\% to 81.8\% accuracy across categories, suggesting reliance on superficial heuristics rather than generalizable aesthetic principles. Notably, these failures persist across model scales: 27B parameter models show no consistent improvement over 8B variants, and reasoning-enhanced LLMs (Claude-4-Opus-thinking, OpenAI-o3) provide no advantage over standard architectures.

\textbf{Contributions.} We make three contributions to understanding preference learning:

\begin{itemize}

\item \textbf{Benchmark}: WritingPreferenceBench provides 1,800 validated preference  pairs with systematic signal isolation across English and Chinese, enabling reproducible cross-lingual evaluation of subjective preference modeling.

\item \textbf{Empirical findings}: Comprehensive evaluation establishes that (i) sequence classifiers fail systematically on subjective tasks, (ii) generative reward models with reasoning achieve 30\% higher accuracy, and (iii) zero-shot LLM judges cannot reliably assess creative quality despite instruction tuning.

\item \textbf{Architectural insights}: Evidence that successful preference learning requires intermediate reasoning representations, not just pattern matching, with implications for next-generation RLHF systems.
\end{itemize}

\section{WritingPreferenceBench}
\label{sec:writingpreferencebench}

\begin{figure*}[!t]
    \centering
    \includegraphics[width=\linewidth]{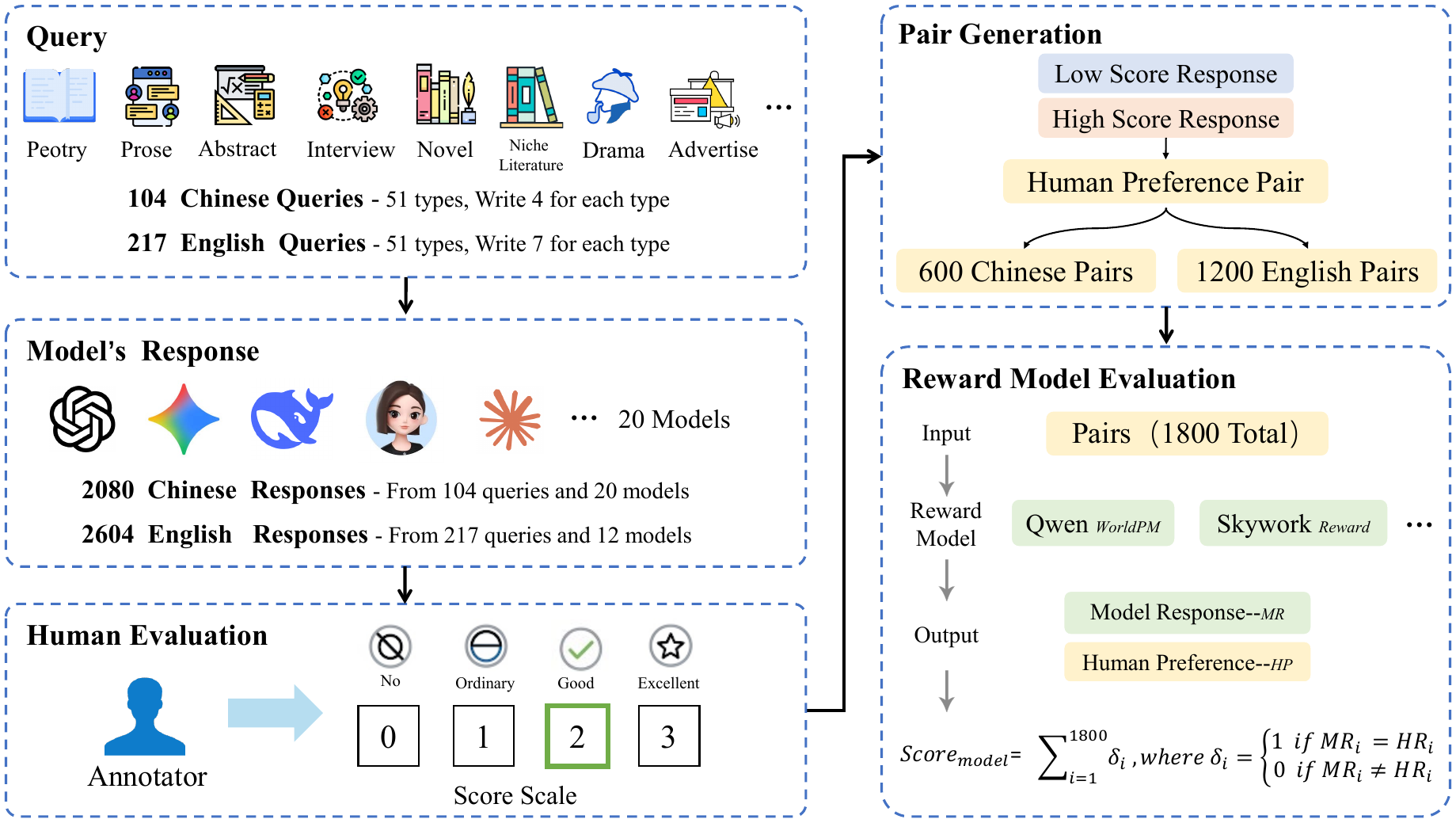} 
    \caption{The data curation pipeline of \textbf{WritingPreferenceBench}. Our multi-stage process begins with expert-crafted queries across 51 genres, generates diverse responses using 20 state-of-the-art models, and culminates in rigorous human evaluation by trained annotators. Quality control mechanisms operate throughout to ensure preference pairs reflect genuine subjective quality distinctions rather than objective differences.}
    \label{fig:data_curation}
\end{figure*}

The fundamental challenge in evaluating subjective writing is not merely collecting human judgments, but ensuring those judgments isolate genuine aesthetic and stylistic preference from objective quality signals. We present \textbf{WritingPreferenceBench}, a meticulously constructed benchmark that addresses this challenge through 1,800 preference pairs spanning English and Chinese creative writing. The construction process, illustrated in Figure \ref{fig:data_curation}, was guided by rigorous design principles and implemented through a human-in-the-loop pipeline designed to systematically eliminate confounding variables.

\subsection{Benchmark Construction Pipeline}

We implemented a multi-stage pipeline that translates our design principles into a concrete, reproducible workflow. This process, depicted in Figure \ref{fig:data_curation}, first generates a diverse and culturally-rich set of candidate responses and then applies a rigorous, human-led filtering protocol to isolate pairs that represent genuine subjective preference.

\textbf{Phase 1: Architecting Diverse and Representative Queries.}
The benchmark's foundation is a taxonomy of 51 creative writing categories, developed by merging taxonomies from established writing communities. To ensure both representational diversity and practical relevance, these categories range from classical literary traditions (e.g., poetry, fiction) to contemporary forms (e.g., advertising copy, social media content). Query development followed a dual-expertise workflow where two experienced creative writing instructors drafted and aligned on creative blueprints for each category. For the English subset, queries were designed to be culturally neutral, avoiding references specific to any single Anglophone culture. For the Chinese subset, queries were adapted to reflect rhetorical conventions and genre expectations common in Mandarin writing traditions, while maintaining structural parallelism with English queries to enable cross-lingual comparison. 

To scale query generation while maintaining quality, we employed a human-AI collaborative approach: the expert-authored blueprints (typically 2-3 sentences outlining the creative task and key constraints) were expanded into full queries using Gemini 2.5 Pro. Each generated query was then independently reviewed by both instructors, who validated it for creative intent, appropriate difficulty, and evaluative granularity. Queries requiring revision underwent iterative refinement (3-5 rounds on average) until both experts reached consensus on quality and alignment with the blueprint specifications.

\textbf{Phase 2: Generating a Spectrum of Responses.}
To create a rich and varied corpus for curation, we utilized a diverse suite of 20 state-of-the-art language models, including GPT-4.1, Claude-4, Gemini-2.5-Pro, and Doubao-1.5-Pro. For each query, every model produced 5 outputs with temperature sampling set to $T=0.8$. This strategy ensured the generation of a wide spectrum of quality—from formulaic to highly original—providing the necessary variance to identify the controlled preference gaps central to our benchmark's design.

\textbf{Phase 3: Human-in-the-Loop Annotation and Quality Control.}
This phase is the cornerstone of our methodology, operationalizing a focus on subjective quality through a rigorous, integrated annotation and filtering protocol.
\begin{itemize}
    \item \textbf{Initial Triage: Filtering for Objective Correctness.} Before subjective assessment, an automated screening process removed responses with objective deficiencies. This filter eliminated approximately 15\% of the raw responses, discarding outputs with comprehension-impeding grammatical errors, factual inconsistencies, or clear prompt violations. This crucial step ensures our benchmark tests for subjective quality, not basic error detection.

    \item \textbf{Expert Evaluation with a Calibrated Rubric.} We recruited 11 expert annotators through a professional annotation service with demonstrated expertise in creative content evaluation. Annotator selection criteria included:
    \begin{itemize}
        \item \textbf{Language proficiency:} English subset annotators (n=4) demonstrated native or near-native fluency; Chinese subset annotators (n=7) were native Mandarin speakers.
        \item \textbf{Writing competency:} All annotators passed a qualification assessment requiring them to critique sample creative texts and provide detailed justifications aligned with our evaluation rubric.
        \item \textbf{Genre familiarity:} Annotators demonstrated knowledge across multiple writing genres through a pre-screening questionnaire covering the 8 macro-categories in our taxonomy.
    \end{itemize}
    
    All annotators completed an 8-hour training program consisting of: (1) instruction on the evaluation rubric and subjective quality dimensions, (2) practice annotation of 50 consensus examples with group discussion of disagreements, and (3) a final calibration phase where inter-annotator agreement was measured. Each annotator then independently scored responses on a 4-point scale designed to distinguish levels of creative quality:
    \begin{itemize}
        \item \textbf{3 (Creative):} Demonstrates genuine creativity, stylistic flair, and deep engagement; suitable for publication.
        \item \textbf{2 (Competent):} Well-structured and complete, but predictable and lacking in originality.
        \item \textbf{1 (Formulaic):} Technically correct but lacks creative engagement; follows a rigid template.
        \item \textbf{0 (Elementary):} Fundamentally flawed or off-topic.
    \end{itemize}
    The framework includes universal quality criteria and detailed genre-specific standards (see Appendix~\ref{appendix:scoring} for complete guidelines).

    \item \textbf{Statistical Validation and Final Pair Curation.} A final preference pair was curated only if it met three strict criteria for reliability and validity. A pair was accepted only if:
    \begin{enumerate}
        \item It had \textbf{directional agreement} from at least 2 of the 3 annotators.
        \item It showed a \textbf{minimum score gap} of $\Delta \ge 1$.
        \item It passed a check for \textbf{absence of confounding factors}, such as significant length disparities.
    \end{enumerate}
    Finally, a separate team of bilingual experts performed cross-lingual validation  on a sample of pairs to confirm that scoring standards were applied equivalently across both languages.
\end{itemize}

\begin{figure}[th]
    \centering
    \includegraphics[width=0.95\textwidth]{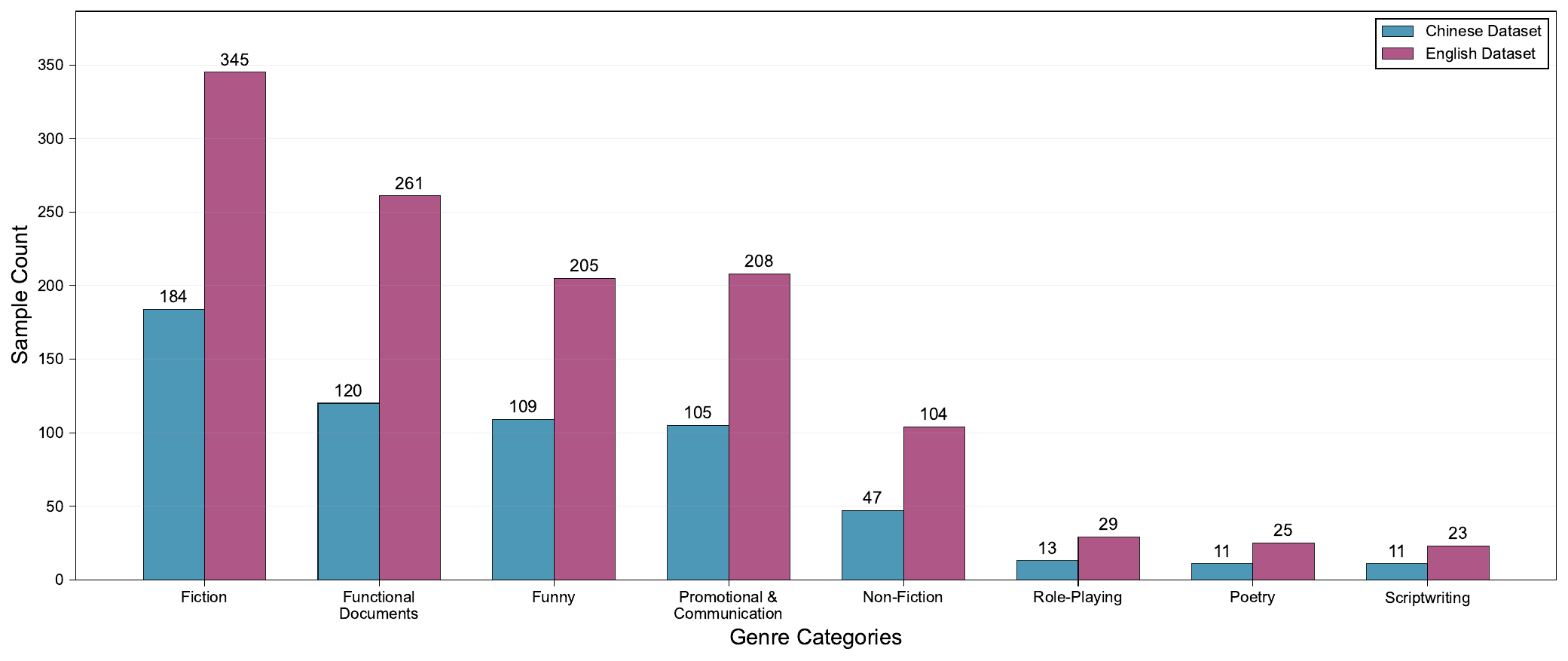}
    \caption{Distribution of preference pairs across a sample of the 8 writing macro-categories for both English and Chinese in \textbf{WritingPreferenceBench}. The dataset maintains balanced coverage across diverse writing genres, with deliberate oversampling of underrepresented categories to ensure comprehensive evaluation of preference modeling capabilities.}
    \label{fig:dist-pairs}
\end{figure}

\subsection{Dataset Statistics}

\textbf{WritingPreferenceBench} comprises 1,800 human-validated preference pairs (1,200 English, 600 Chinese) that establish a new standard for evaluating subjective writing preferences. Unlike existing benchmarks that conflate objective correctness with aesthetic quality, our dataset isolates genuine stylistic preference through careful statistical design.

\textbf{Compositional Structure and Coverage.} 

Figure~\ref{fig:dist-pairs} reveals the deliberate architectural choices underlying our 51-category taxonomy. The dataset's statistical properties reflect three critical design decisions:

\begin{itemize}
    \item \textbf{Cross-lingual parity}: While maintaining a 2:1 English-Chinese ratio due to annotator availability, we ensure equivalent statistical power across languages with a minimum of 20 pairs per category in each language.
    
    \item \textbf{Genre equilibrium}: Each category contains 20-40 preference pairs (mean=35.3, std=7.2), a distribution engineered to prevent the genre collapse observed in web-scraped datasets. 
    
    \item \textbf{Compositional diversity}: The taxonomy spans 8 macro-categories with deliberate oversampling of traditionally underrepresented genres (e.g., poetry, scriptwriting) to stress-test models' preference modeling capabilities beyond dominant web text distributions.
\end{itemize}

\begin{figure*}[t]
    \centering
    \begin{subfigure}[b]{0.48\textwidth}
        \centering
        \includegraphics[width=\textwidth]{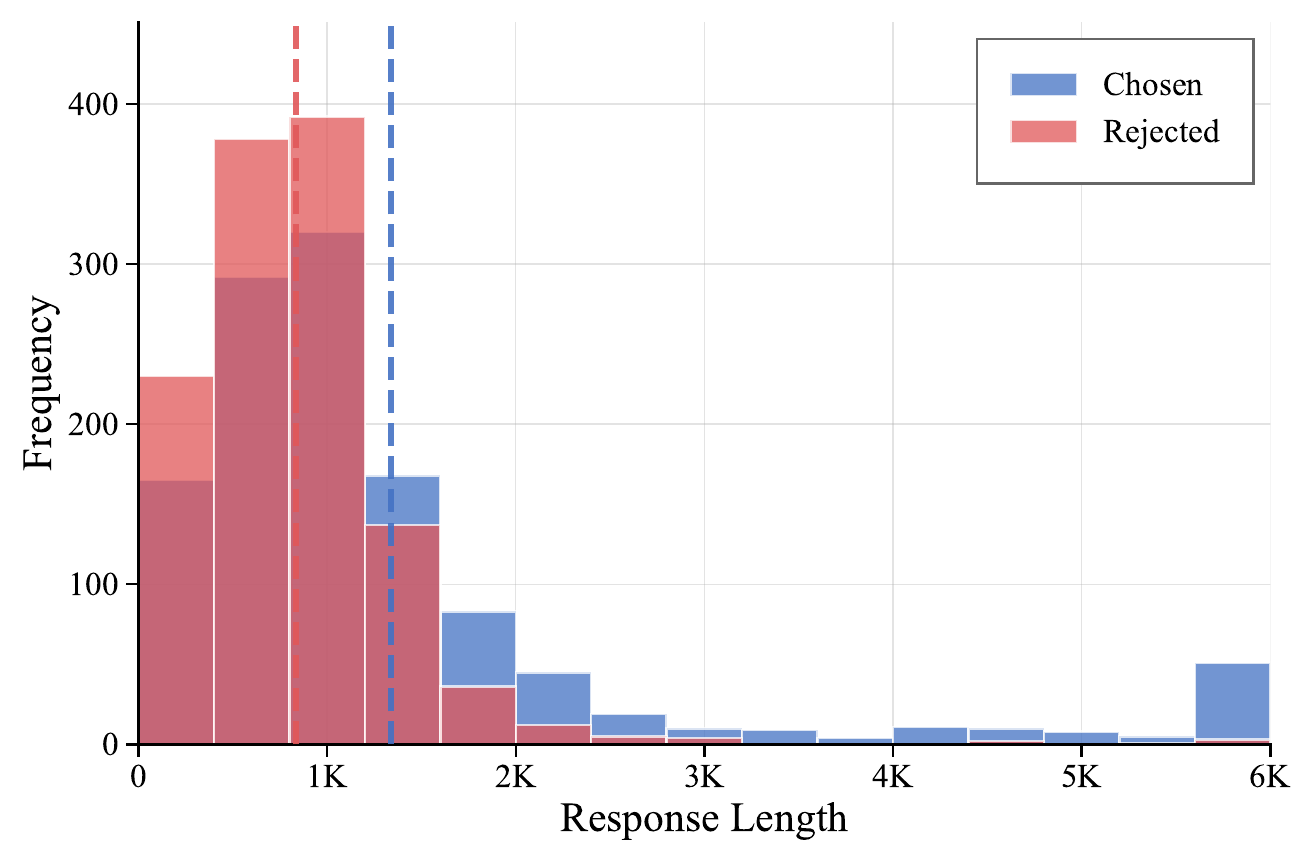}
        \caption{English dataset}
        \label{fig:wpb_english_sub}
    \end{subfigure}
    \hfill
    \begin{subfigure}[b]{0.48\textwidth}
        \centering
        \includegraphics[width=\textwidth]{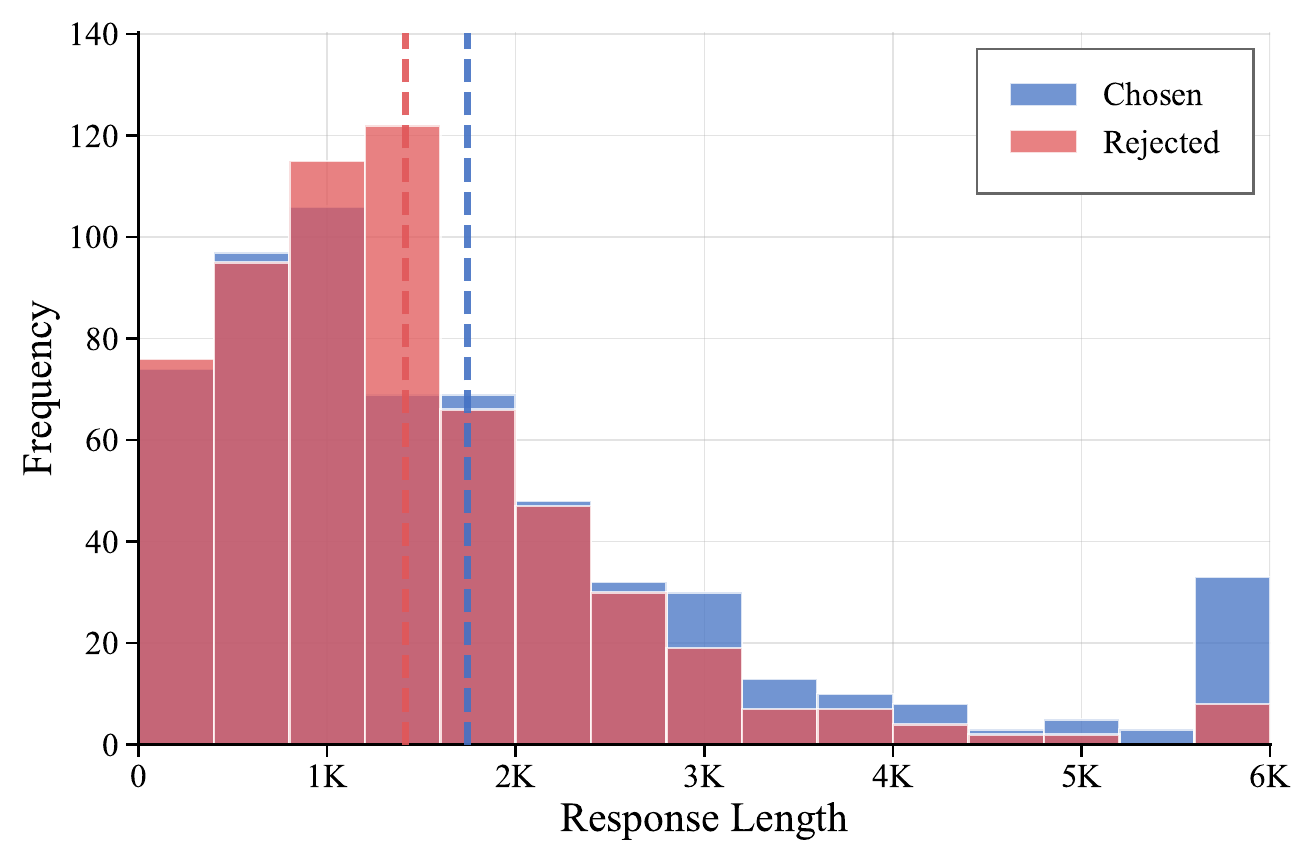}
        \caption{Chinese dataset}
        \label{fig:wpb_chinese_sub}
    \end{subfigure}
    \caption{Length distributions of chosen and rejected responses reveal the statistical signature of creative quality in \textbf{WritingPreferenceBench}. Distributions truncated at 6K for visualization; dashed lines indicate means.}
    \label{fig:wpb_distributions}
\end{figure*}

\textbf{Statistical Validation of Subjective Quality Gaps.}
Figure~\ref{fig:wpb_distributions} and Table~\ref{tab:dataset_stats} reveal the empirical signature of subjective preference in our dataset. The length distributions expose a critical phenomenon: chosen responses exhibit significantly higher variance (English: SD=1801.9 vs. 593.4; Chinese: SD=1967.5 vs. 1311.0) and right-skewness compared to rejected responses. This asymmetry reflects a fundamental property of creative excellence—while mediocrity converges toward formulaic patterns, creativity manifests across diverse scales.
The score distributions validate our annotation protocol's effectiveness. The median scores (English: 3 vs. 2; Chinese: 3 vs. 1) align precisely with our rubric's creative-competent and creative-formulaic boundaries, demonstrating that our benchmark captures the most informative preference contrasts.

\section{Experiments}

We evaluate 21 models on WritingPreferenceBench: 7 reward models and 14 language models serving as zero-shot judges. This section describes our evaluation protocols and experimental setup.

\subsection{Evaluation Protocols}

\textbf{Protocol 1: Reward Model Scoring.} 
For each preference pair $(R_{\text{chosen}}, R_{\text{rejected}})$, reward models assign scalar scores. A prediction is correct if $\text{RM}(R_{\text{chosen}}) > \text{RM}(R_{\text{rejected}})$. We compute accuracy as:
$$\text{Accuracy} = \frac{1}{N}\sum_{i=1}^{N} \mathbb{I}[\text{RM}(R^{(i)}_{\text{chosen}}) > \text{RM}(R^{(i)}_{\text{rejected}})]$$
where $N$ denotes total preference pairs and $\mathbb{I}[\cdot]$ is the indicator function.

\textbf{Protocol 2: Pairwise Preference Judgment.} 
Language models receive both responses with instructions to select the preferred text based on creativity, style, and emotional resonance. We use deterministic decoding ($T=0$) and extract preferences from model outputs. This protocol tests whether general-purpose models can perform zero-shot preference evaluation without specialized training.

\subsection{Models}

\textbf{Reward Models.} 
We evaluate 7 models spanning different architectures and scales:
\begin{itemize}
\item \textbf{Sequence Classifiers}: Nvidia/AceMath-7B-RM~\citep{acemath2024}, RM-Mistral-7B~\citep{dong2023raft}, Skywork-Reward-Llama-3.1-8B~\citep{liu2024skywork}, Skywork-Reward-Gemma-2-27B~\cite{liu2024skywork}
\item \textbf{Generative RMs}: RM-R1-DeepSeek-Qwen-7B, RM-R1-DeepSeek-Qwen-14B, RM-R1-Qwen2.5-7B~\citep{chen2025rm,qwen2.5,guo2025deepseek}
\end{itemize}

\textbf{Language Model Judges.} 
We evaluate 14 models including reasoning-enhanced variants (Claude-4-\{Opus, Sonnet\}-thinking~\citep{anthropic2025claude4}, Doubao-1.6-thinking~\citep{seed2025seed1}, OpenAI-o3-high~\citep{openai2025o3high}) and standard models (Gemini-2.5-\{Flash, Pro\}~\citep{geminiteam2024gemini15unlockingmultimodal}, DeepSeek-R1~\citep{guo2025deepseek}, ByteDance-Seed-1.6~\citep{bytedance2025seed1.6}, Doubao-\{1.5-Lite, 1.5-Pro, 1.6-flash\}~\citep{bytedance_doubao15_pro_2025}, Qwen-3-235B~\citep{qwen2.5}, OpenAI-o4-mini).

\subsection{Implementation Details}

All experiments use the same prompt templates across models to ensure fair comparison. For reward models, we use the default inference configuration from their respective repositories. For LLM judges, we employ a standardized prompt format that presents both responses and requests a preference judgment with justification. We evaluate on the full WritingPreferenceBench dataset (1,200 English, 600 Chinese pairs) without subsampling.

\section{Results}

\subsection{Reward Model Performance}

We evaluate seven reward models across WritingPreferenceBench. Following RewardBench~\citep{lambert2024rewardbench}, models divide into sequence classifiers (discriminative heads on language models) and generative reward models. However, our results reveal that the RM-R1 series~\citep{chen2025rm} represents a distinct category—generative models that produce reasoning chains before preference judgments, diverging from both traditional classifiers and DPO-based approaches evaluated in RewardBench.
Table~\ref{tab:rm_results} presents accuracy across eight writing genres in English and Chinese.

\paragraph{Sequence classifiers fail on subjective preference.}
Traditional sequence-based reward models achieve 52.7\% mean accuracy across both languages (ranging from 46.8\% to 62.6\% on individual language subsets), statistically indistinguishable from random chance (binomial test, $p>0.05$). This aggregate masks severe instability: all four models exhibit catastrophic failures on at least one genre, with Nvidia/AceMath-7B dropping to 18.2\% on Chinese Poetry while reaching 61.5\% on Role-Playing—a 43.3 percentage point swing within a single model. Three of four sequence classifiers fall below 50\% on their weaker language, indicating systematic rather than random failures.

\paragraph{Generative architecture enables strong performance.}
Generative reward models achieve substantially higher accuracy than sequence classifiers. RM-R1-Qwen2.5-7B reaches 81.8\% (EN)—the highest performance across all models and 30 percentage points above the sequence classifier mean. All three generative models exceed 50\% accuracy, while three of four sequence classifiers fall below random chance. This architectural advantage persists across languages: best generative performance reaches 64.5\% (ZH) versus 53.5\% for sequence classifiers.

\paragraph{Scale improves stability, not just accuracy.}
Scaling from 7B to 14B in RM-R1-DeepSeek yields distinct benefits: accuracy improves (50.3\%→62.6\% ZH), but more critically, variance drops (9.8→5.5). This stability gain does not transfer to sequence classifiers—Skywork-Gemma-27B shows no improvement over 8B variants despite 3.4× parameters. The 14B model's low variance (5.5) represents the most consistent performance across genres, suggesting scale enables robust preference representations in generative architectures.

\begin{table}[t]
\centering
\caption{Reward model accuracy (\%) on \textbf{WritingPreferenceBench} by architecture. Colors: \colorbox{green!15}{$>$70\%}, \colorbox{red!15}{$<$50\%}. Best overall in \textbf{bold}.}
\label{tab:rm_results}
\resizebox{\textwidth}{!}{%
\begin{tabular}{@{}ll|cccccccc|cc@{}}
\toprule
\textbf{Model} & \textbf{Lang} & \textbf{Func.} & \textbf{Promo.} & \textbf{Non-Fic.} & \textbf{Fiction} & \textbf{Funny} & \textbf{Poetry} & \textbf{Script} & \textbf{Role} & \textbf{Avg} & \textbf{Std} \\
\midrule
\multicolumn{12}{c}{\textit{\textbf{Sequence Classifiers} (Discriminative with scalar output)}} \\
\midrule
\multirow{2}{*}{Nvidia/AceMath-7B} 
& ZH & 56.7 & 50.5 & 55.3 & 54.9 & 52.3 & \cellcolor{red!15}18.2 & 54.6 & 61.5 & 53.5 & 11.2 \\
& EN & \cellcolor{red!15}48.0 & 53.9 & 59.6 & \cellcolor{red!15}33.9 & 55.1 & \cellcolor{red!15}36.0 & \cellcolor{red!15}21.7 & 51.7 & \cellcolor{red!15}46.8 & 12.4 \\
\cmidrule{1-12}
\multirow{2}{*}{RM-Mistral-7B} 
& ZH & 60.8 & \cellcolor{red!15}46.7 & 63.8 &55.9 & 54.1 &54.5 & \cellcolor{green!15}72.7 & \cellcolor{red!15}46.1 & 55.6 & 9.1 \\
& EN & 65.2 & 60.0 & 54.8 & 62.9 & 64.9 & \cellcolor{green!15}72.0 & \cellcolor{green!15}78.3 & \cellcolor{red!15}44.8 & 62.6 & 10.1 \\
\cmidrule{1-12}
\multirow{2}{*}{Skywork-Llama-3.1-8B} 
& ZH & 56.7 & \cellcolor{red!15}43.8 & 61.7 & 52.2 & 50.5 & 63.6 & 54.6 & \cellcolor{red!15}38.5 & 52.0 & 8.2 \\
& EN & 53.6 & 56.3 & 60.6 & \cellcolor{red!15}49.0 & 52.2 & 56.0 & 65.2 & \cellcolor{red!15}41.4 & 53.1 & 7.3 \\
\cmidrule{1-12}
\multirow{2}{*}{Skywork-Gemma-2-27B} 
& ZH & 50.8 & 54.3 & 55.3 & 53.3 & \cellcolor{red!15}40.4 & \cellcolor{green!15}81.8 & \cellcolor{red!15}45.5 & 53.9 & 51.2 & 11.3 \\
& EN & \cellcolor{red!15}49.0 & 53.9 & 59.6 & \cellcolor{red!15}33.9 & 55.1 & \cellcolor{red!15}36.0 & \cellcolor{red!15}21.7 & 51.7 & \cellcolor{red!15}46.8 & 12.4 \\
\midrule
\multicolumn{12}{c}{\textit{\textbf{Generative Reward Models} (Reasoning before scoring)}} \\
\midrule
\multirow{2}{*}{RM-R1-DeepSeek-Qwen-7B} 
& ZH & 50.8 & \cellcolor{red!15}45.7 & 57.5 & 56.5 & \cellcolor{red!15}45.0 & \cellcolor{red!15}27.3 & \cellcolor{red!15}36.4 & \cellcolor{red!15}46.2 & 50.3 & 9.8 \\
& EN & 64.8 & 50.9 & 58.7 & 57.4 & 54.4 & 52.0 & 52.2 & \cellcolor{red!15}37.9 & 56.8 & 7.2 \\
\cmidrule{1-12}
\multirow{2}{*}{RM-R1-DeepSeek-Qwen-14B} 
& ZH & 59.2 & \cellcolor{red!15}45.7 & 57.5 & \cellcolor{green!15}71.7 & 66.7 & \cellcolor{green!15}90.0 & 54.6 & 69.2 & {62.6} & 13.2 \\
& EN & \cellcolor{green!15}71.3 & 61.5 & 63.5 & 58.6 & 59.0 & 64.0 & 52.2 & 65.5 & 62.5 & 5.5 \\
\cmidrule{1-12}
\multirow{2}{*}{RM-R1-Qwen2.5-7B} 
& ZH & 69.1& 56.1 & \cellcolor{green!15}78.7 & \cellcolor{green!15}86.4 & 67.9 & \cellcolor{green!15}81.8 & \cellcolor{green!15}72.7 & \cellcolor{green!15}84.6 & \cellcolor{green!15}\textbf{73.3}& 10.9 \\
& EN & \cellcolor{green!15}79.3 & \cellcolor{green!15}76.0 & \cellcolor{green!15}91.4 & \cellcolor{green!15}89.6 & \cellcolor{green!15}72.9 & \cellcolor{green!15}92.0 & \cellcolor{green!15}85.9 & 65.5 & \cellcolor{green!15}\textbf{81.8} & 9.5 \\
\bottomrule
\end{tabular}
}\cellcolor{red!15}
\end{table}

\paragraph{Sequence classifiers exhibit catastrophic genre failure.}
All sequence classifiers demonstrate extreme performance swings: Nvidia/AceMath-7B 
ranges from 18.2\% to 61.5\% (43.3 percentage point gap), while Skywork-Gemma-27B 
varies from 21.7\% to 81.8\%. Mean within-model standard deviation reaches 10.1\% 
for discriminative models versus 9.2\% for generative. 
These genre-specific failures—often below 40\% accuracy—indicate fundamental 
instability rather than minor variations.

\textbf{Cross-lingual consistency reveals architectural robustness.}
Generative models maintain more consistent cross-lingual performance than sequence classifiers. RM-R1-DeepSeek-Qwen-14B achieves 62.6\% (ZH) and 62.5\% (EN), while sequence classifiers show larger gaps: Nvidia/AceMath-7B scores 53.5\% (ZH) versus 46.8\% (EN). This consistency in generative models, particularly at larger scales, suggests that reasoning-based architectures learn more language-agnostic preference representations.

\subsection{Language Model Judge Performance}

Table~\ref{tab:llm_judge_results} presents the performance of 14 state-of-the-art language models serving as zero-shot preference judges on WritingPreferenceBench, revealing systematic underperformance compared to specialized reward models.

\begin{table}[!ht]
\centering
\caption{Language model judge accuracy (\%) on \textbf{WritingPreferenceBench} using pairwise preference evaluation. Colors: \colorbox{green!15}{$>$70\%}, \colorbox{red!15}{$<$50\%}. Best overall in \textbf{bold}.}
\label{tab:llm_judge_results}
\resizebox{\textwidth}{!}{%
\begin{tabular}{@{}ll|cccccccc|cc@{}}
\toprule
\textbf{Model} & \textbf{Lang} & \textbf{Func.} & \textbf{Promo.} & \textbf{Non-Fic.} & \textbf{Fiction} & \textbf{Funny} & \textbf{Poetry} & \textbf{Script} & \textbf{Role} & \textbf{Avg} & \textbf{Std} \\
\midrule
\multirow{2}{*}{ByteDance-Seed-1.6} & ZH & 42.1 & 32.2 & 52.5 & 59.9 & 38.5 & \cellcolor{red!15}45.5 & \cellcolor{red!15}36.4 & \cellcolor{red!15}38.5 & \cellcolor{red!15}45.5 & 9.4 \\
& EN & 54.6 & 54.2 & \cellcolor{red!15}49.2 & \cellcolor{red!15}41.1 & \cellcolor{red!15}47.3 & 68.0 & \cellcolor{red!15}17.4 & \cellcolor{red!15}41.4 & \cellcolor{red!15}48.3 & 13.4 \\
\cmidrule{1-12}
\multirow{2}{*}{Claude-4-Opus-thinking} & ZH & 55.1 & \cellcolor{red!15}36.4 & 57.6 & \cellcolor{green!15}73.3 & \cellcolor{red!15}49.5 & 54.6 & \cellcolor{green!15}72.7 & \cellcolor{red!15}46.2 & 56.0 & 12.1 \\
& EN & 65.7 & 64.3 & 64.1 & 60.1 & 54.2 & 64.0 & \cellcolor{red!15}43.5 & 51.7 & 61.0 & 7.3 \\
\cmidrule{1-12}
\multirow{2}{*}{Claude-4-Sonnet-thinking} & ZH & \cellcolor{red!15}46.7 & \cellcolor{red!15}38.1 & 62.7 & 65.7 & \cellcolor{red!15}48.6 & 54.6 & 63.6 & \cellcolor{red!15}46.2 & 52.8 & 9.9 \\
& EN & 62.4 & 58.6 & 58.7 & 53.9 & 50.3 & 50.0 & \cellcolor{red!15}31.8 & \cellcolor{red!15}48.2 & 55.7 & 9.3 \\
\cmidrule{1-12}
\multirow{2}{*}{DeepSeek-R1} & ZH & \cellcolor{red!15}46.7 & \cellcolor{red!15}41.5 & 61.0 & 61.6 & \cellcolor{red!15}48.6 & \cellcolor{red!15}45.5 & 63.6 & \cellcolor{red!15}46.2 & 52.0 & 8.8 \\
& EN & 57.4 & 61.7 & \cellcolor{red!15}43.8 & \cellcolor{red!15}40.8 & \cellcolor{red!15}42.9 & \cellcolor{green!15}72.0 & \cellcolor{red!15}17.4 & 51.7 & \cellcolor{red!15}49.3 & 15.4 \\
\cmidrule{1-12}
\multirow{2}{*}{Doubao-1.5-Lite} & ZH & \cellcolor{red!15}44.9 & \cellcolor{red!15}42.4 & 64.4 & 62.8 & \cellcolor{red!15}44.0 & \cellcolor{red!15}36.4 & \cellcolor{green!15}72.7 & \cellcolor{red!15}38.5 & 51.5 & 13.0 \\
& EN & 62.8 & 63.9 & \cellcolor{red!15}42.2 & 50.5 & \cellcolor{red!15}45.4 & 52.0 & \cellcolor{red!15}47.8 & \cellcolor{red!15}48.3 & 53.7 & 8.1 \\
\cmidrule{1-12}
\multirow{2}{*}{Doubao-1.5-Pro} & ZH & 54.2 & \cellcolor{red!15}47.5 & \cellcolor{green!15}72.9 & \cellcolor{green!15}79.7 & 54.1 & 54.6 & 63.6 & 69.2 & 62.5 & 10.8 \\
& EN & \cellcolor{green!15}74.4 & \cellcolor{green!15}74.5 & 64.8 & \cellcolor{green!15}71.0 & 57.1 & 68.0 & 56.5 & 58.6 & \cellcolor{green!15}\textbf{68.7} & 7.2 \\
\cmidrule{1-12}
\multirow{2}{*}{Doubao-1.6-flash} & ZH & \cellcolor{red!15}39.3 & \cellcolor{red!15}30.5 & 55.9 & 60.5 & \cellcolor{red!15}38.5 & 54.6 & 63.6 & \cellcolor{red!15}38.5 & \cellcolor{red!15}45.8 & 11.9 \\
& EN & 57.0 & 53.7 & \cellcolor{red!15}49.2 & \cellcolor{red!15}43.6 & \cellcolor{red!15}46.3 & 52.0 & \cellcolor{red!15}30.4 & \cellcolor{red!15}44.8 & \cellcolor{red!15}49.3 & 7.7 \\
\cmidrule{1-12}
\multirow{2}{*}{Doubao-1.6-thinking} & ZH & \cellcolor{red!15}46.7 & \cellcolor{red!15}35.6 & 66.1 & \cellcolor{green!15}73.8 & \cellcolor{red!15}43.1 & \cellcolor{green!15}72.7 & 54.6 & \cellcolor{red!15}46.2 & 54.2 & 13.9 \\
& EN & 64.1 & 66.5 & 60.9 & 57.9 & \cellcolor{red!15}48.8 & \cellcolor{green!15}72.0 & \cellcolor{red!15}30.4 & \cellcolor{red!15}41.4 & 58.9 & 12.8 \\
\cmidrule{1-12}
\multirow{2}{*}{Doubao-1.6-thinking-agent} & ZH & \cellcolor{red!15}48.6 & \cellcolor{red!15}40.7 & 66.1 & \cellcolor{green!15}71.5 & \cellcolor{red!15}49.5 & 63.6 & 63.6 & 53.9 & 56.2 & 10.3 \\
& EN & 64.9 & 61.7 & 61.7 & 55.5 & \cellcolor{red!15}47.3 & 68.0 & \cellcolor{red!15}39.1 & \cellcolor{red!15}48.3 & 57.6 & 9.6 \\
\cmidrule{1-12}
\multirow{2}{*}{Gemini-2.5-Flash} & ZH & \cellcolor{red!15}47.7 & \cellcolor{red!15}34.8 & 61.0 & 68.0 & \cellcolor{red!15}45.0 & 63.6 & 63.6 & \cellcolor{red!15}38.5 & 52.2 & 12.1 \\
& EN & 59.1 & 57.7 & 62.5 & 59.8 & 52.2 & 56.0 & \cellcolor{red!15}34.8 & 51.7 & 57.5 & 8.1 \\
\cmidrule{1-12}
\multirow{2}{*}{Gemini-2.5-Pro} & ZH & 53.3 & \cellcolor{red!15}44.9 & \cellcolor{green!15}71.2 & \cellcolor{green!15}\textbf{80.2} & \textbf{56.9} & \cellcolor{green!15}\textbf{72.7} & \cellcolor{green!15}72.7 & 61.5 & \textbf{62.7} & 11.3 \\
& EN & \cellcolor{green!15}70.3 & 66.5 & \cellcolor{green!15}\textbf{68.0} & 65.7 & \textbf{58.5} & \cellcolor{green!15}\textbf{80.0} & \cellcolor{red!15}34.8 & \cellcolor{green!15}\textbf{72.4} & 65.7 & 12.6 \\
\cmidrule{1-12}
\multirow{2}{*}{OpenAI-o4-mini} & ZH & \cellcolor{red!15}43.0 & \cellcolor{red!15}33.9 & 54.2 & 50.6 & \cellcolor{red!15}35.8 & \cellcolor{red!15}45.5 & 63.6 & \cellcolor{red!15}38.5 & \cellcolor{red!15}43.5 & 9.9 \\
& EN & 58.3 & 58.6 & 60.9 & 55.5 & 53.2 & 68.0 & \cellcolor{red!15}30.4 & 55.2 & 56.6 & 10.0 \\
\cmidrule{1-12}
\multirow{2}{*}{OpenAI-o3-high} & ZH & \cellcolor{red!15}36.5 & \cellcolor{red!15}28.0 & 52.5 & 50.6 & \cellcolor{red!15}40.4 & 63.6 & 54.6 & \cellcolor{red!15}38.5 & \cellcolor{red!15}42.0 & 11.1 \\
& EN & 55.4 & \cellcolor{red!15}45.8 & 54.7 & \cellcolor{red!15}46.7 & \cellcolor{red!15}41.0 & \cellcolor{green!15}72.0 & \cellcolor{red!15}21.7 & \cellcolor{red!15}41.4 & \cellcolor{red!15}48.1 & 14.0 \\
\cmidrule{1-12}
\multirow{2}{*}{Qwen-3-235B} & ZH & \cellcolor{red!15}45.8 & \cellcolor{red!15}39.0 & 59.3 & 62.2 & \cellcolor{red!15}42.2 & 54.6 & 63.6 & 53.9 & 50.5 & 9.2 \\
& EN & 57.9 & \cellcolor{red!15}46.7 & \cellcolor{red!15}45.3 & \cellcolor{red!15}43.0 & \cellcolor{red!15}40.5 & 64.0 & \cellcolor{red!15}17.4 & \cellcolor{red!15}41.4 & \cellcolor{red!15}46.4 & 13.3 \\
\bottomrule
\end{tabular}
}
\end{table}

\textbf{LLM judges systematically underperform reward models.}
General-purpose language models achieve mean accuracy of 53.9\%, compared to 58.2\% for reward models—a 4.3\% degradation despite orders of magnitude more parameters. The best LLM judge (Doubao-1.5-Pro: 68.7\% EN) remains 13.1\% below the top generative reward model (RM-R1-Qwen2.5-7B: 81.8\% EN). This gap persists across all model families and scales, indicating that zero-shot preference evaluation cannot match task-specific training.

\textbf{Reasoning capabilities provide no systematic advantage.}
Models with explicit reasoning mechanisms show no consistent improvement over standard architectures. Claude-4-Opus-thinking achieves 61.0\% (EN) while non-reasoning Doubao-1.5-Pro reaches 68.7\%. Similarly, OpenAI-o3-high with advanced reasoning scores only 48.1\%, performing worse than simpler models like Gemini-2.5-Flash (57.5\%). The correlation between reasoning capability and preference accuracy is negligible (r=0.08, p>0.5), suggesting that chain-of-thought processing does not inherently improve subjective quality assessment.

\textbf{Genre instability exceeds that of reward models.}
LLM judges exhibit extreme performance variance across genres, surpassing even sequence classifiers. Gemini-2.5-Pro ranges from 80.0\% on English Poetry to 34.8\% on Scriptwriting—a 45.2\% gap. OpenAI-o3-high shows similar instability: 72.0\% on Poetry versus 21.7\% on Scriptwriting. Mean within-model standard deviation reaches 11.4\%, with 9 of 14 models showing standard deviations exceeding 10\%. This variance pattern suggests that LLMs rely on superficial genre markers rather than genuine quality assessment.

\textbf{Cross-lingual performance reveals model-specific biases.}
LLM judges demonstrate inconsistent cross-lingual patterns. Doubao models maintain relative consistency (1.5-Pro: 62.5\% ZH, 68.7\% EN), while others show severe degradation: OpenAI-o3-high drops from 48.1\% (EN) to 42.0\% (ZH). These disparities do not correlate with known multilingual capabilities, suggesting that preference evaluation activates different model behaviors across languages.

\textbf{Implications for LLM-as-judge paradigm.}
The systematic underperformance of LLM judges relative to specialized reward models challenges the widespread adoption of LLM-as-judge evaluation~\citep{zheng2023judging}. Mean accuracy of 53.9\%—barely above random—indicates that zero-shot prompting cannot elicit reliable preference judgments for subjective tasks. The failure of reasoning-enhanced models further suggests that the limitation is not computational but representational: without explicit preference training, even advanced LLMs default to surface-level heuristics rather than genuine quality assessment.

\section{Discussion}
Our findings reveal fundamental limitations in current preference learning paradigms when applied to subjective domains and expose a critical gap between model capabilities and genuine human aesthetic judgment.

\textbf{A Performance Ceiling on Subjectivity.} Even the best-performing models struggle to surpass a modest accuracy threshold on purely subjective tasks. The top bilingual reward model achieves only ~62.5\% accuracy, suggesting that current methods are more adept at identifying objective flaws (which we filtered out) than they are at capturing nuanced stylistic and creative preferences. This indicates a potential ceiling for architectures trained primarily on objective-centric data.

\textbf{Explicit Reasoning is No Panacea.} The systematic underperformance of advanced LLM judges compared to simpler, specialized reward models is a striking result. It challenges the prevailing assumption that enhanced reasoning capabilities, such as chain-of-thought, naturally lead to better alignment with complex human values. Our results suggest the problem is not one of logic but of representation; models lack the underlying framework to encode and weigh aesthetic qualities like originality, emotional resonance, or stylistic flair.

\textbf{Preference Functions are Brittle and Unstable.} The most concerning discovery is the extreme performance variance of individual models across genres. Swings of over 50 percentage points between categories like Poetry and Scriptwriting reveal that models are not learning generalizable principles of "good writing." Instead, they appear to be memorizing brittle, genre-specific heuristics. This instability has profound implications for RLHF, as optimizing against such a volatile reward signal could introduce unpredictable and undesirable biases into model behavior, rewarding stylistic mimicry over genuine quality.

\textbf{Cross-Lingual Performance Reflects Training Data Imbalances.} Our cross-lingual  analysis found no consistent patterns in performance differences between English and  Chinese across models. Instead, the gaps appear to be model-specific artifacts, likely stemming from imbalances in training corpora rather than systematic differences in  linguistic structure or aesthetic principles. For instance, RM-R1-Qwen2.5-7B achieves  81.8\% on English but 73.3\% on Chinese, while Nvidia/AceMath-7B shows the reverse  pattern (53.5\% Chinese, 46.8\% English). This variability suggests that current models do not learn language-invariant preference representations, but instead encode  language-specific biases from their pre-training and fine-tuning data. Achieving robust  cross-lingual preference modeling will require training approaches that explicitly  encourage language-agnostic aesthetic understanding.

\section{Related Work}

\textbf{Preference Learning and Evaluation Benchmarks.}
Modern preference learning originated with Christiano et al.~\citep{christiano2017deep}, scaled through InstructGPT~\citep{ouyang2022training} and RLHF~\citep{christiano2017deep}. Subsequent benchmarks~\citep{gao2023scaling,bai2022constitutional,rafailov2023direct,chiang2024chatbot} and comprehensive evaluations like RewardBench~\citep{lambert2024rewardbench}, AlpacaEval~\citep{alpaca_eval}, and MT-Bench~\citep{zheng2023judging} measure conversation, reasoning, and instruction-following. However, these benchmarks conflate objective correctness with subjective preference. RewardBench achieves 95\% on safety but cannot evaluate aesthetic judgment; MT-Bench measures factual accuracy, not creative quality. Our work reveals that models excelling on these benchmarks fail catastrophically (52.7\% accuracy) when objective signals are neutralized.

\textbf{Evaluating Creative and Subjective Writing.}
Creative writing evaluation faces inherent subjectivity challenges~\citep{chodorow2007detection,burstein2003finding,miltsakaki2000automated,li2018coherence,tay2018skipflow}. Early reference-based metrics (BLEU, ROUGE) fail for open-ended generation. Recent benchmarks make progress but retain critical limitations. LitBench~\citep{fein2025litbenchbenchmarkdatasetreliable} uses Reddit upvotes—confounding preference with popularity and timing—and covers only English. WritingBench~\citep{wu2025writingbench} spans six domains but mixes subjective creative tasks with objective functional ones (Academic \& Engineering, Finance \& Business). AlignBench~\citep{liu2023alignbench} evaluates Chinese LLM alignment but focuses on general capabilities rather than creative preference. WritingPreferenceBench advances beyond these by: (1) systematically neutralizing objective confounds through human validation, (2) providing cross-lingual coverage with consistent methodology, and (3) isolating purely subjective quality discrimination where prior benchmarks conflate multiple signals.

\section{Conclusion}

We introduced \textbf{WritingPreferenceBench}, a benchmark isolating subjective writing preference through systematic neutralization of objective quality signals. Our empirical evaluation demonstrates that sequence-based reward models—the dominant RLHF architecture—achieve 52.7\% accuracy on subjective preference tasks. In contrast, generative reward models incorporating explicit reasoning achieve 81.8\% accuracy, suggesting that intermediate representations are necessary for subjective quality assessment. All evaluated models exhibit high variance across genres ($\sigma$=10.1-14.0\%), with individual models ranging from 18.2\% to 92\% accuracy across categories, indicating reliance on genre-specific heuristics rather than generalizable preference functions.

These results have theoretical and practical implications for preference learning. The 30 percentage point performance gap between architectures challenges the direct preference optimization paradigm~\citep{rafailov2023direct} and suggests that subjective domains require fundamentally different inductive biases than objective tasks. The failure of scale to improve performance (27B models underperform 7B variants) and the inability of reasoning-enhanced LLMs to surpass task-specific training indicate that current scaling laws may not apply to subjective preference modeling. Future work should investigate hybrid architectures combining the computational efficiency of discriminative models with the representational capacity of generative reasoning, and develop training objectives that explicitly encourage genre-invariant preference learning.

\clearpage

\input{sections/author_list}

\bibliographystyle{plainnat}
\bibliography{main}
\newpage

\appendix
\input{sections/appendixA}
\end{document}

%% file: math_commands.tex
\usepackage{amsmath,amsfonts,bm}
\usepackage{multicol}

\def\eqref#1{equation~\ref{#1}}

\def\1{\bm{1}}

\DeclareMathAlphabet{\mathsfit}{\encodingdefault}{\sfdefault}{m}{sl}
\SetMathAlphabet{\mathsfit}{bold}{\encodingdefault}{\sfdefault}{bx}{n}



%% file: sections/author_list.tex
\section*{Contributions and Acknowledgements}

Multimodal Art Projection (M-A-P) is a non-profit open-source AI research community, ran by donation.
The community members are working on research topics in a wide range of spectrum, including but not limited to the pre-training paradigm of foundation models, large-scale data collection and processing, and the derived applications on coding, reasoning and music generation.

\textbf{Core Contributors (Equal Contribution)}

\begin{itemize}
    \item Shuangshuang Ying, M-A-P
    \item Yunwen Li, CUHK-Shenzhen and M-A-P
    \item Xingwei Qu, ByteDance Seed, China and The University of Manchester
\end{itemize}

\textbf{Contributors}
\begin{itemize}
    \item Xin Li, Nanyang Technological University
    \item Sheng Jin, Zhejiang University
    \item Minghao Liu, 2077AI and M-A-P
    \item Zhoufutu Wen, ByteDance Seed, China
    \item Xeron Du, M-A-P
    \item Tianyu Zheng, M-A-P
    \item Yichi Zhang, Nanyang Technological University
    \item Letian Ni, Beijing University of Posts and Telecommunications
    \item Yuyang Cheng, Peking University
    \item Zhenzhu Yang, M-A-P
    \item Qiguang Chen, Harbin Institute of Technology
    \item Jingzhe Ding, ByteDance Seed, China
    \item Shengda Long, ByteDance Seed, China
    \item Wangchunshu Zhou, OPPO
    \item Jiazhan Feng, ByteDance Seed, China
    \item Wanjun Zhong, ByteDance Seed, China
\end{itemize}

\textbf{Advisors}
    \begin{itemize}
        \item Libo Qin, Central South University
        \item Wenhao Huang, ByteDance Seed, China
        \item Wanxiang Che, Harbin Institute of Technology
        \item Chenghua Lin, The University of Manchester     
    \end{itemize}

\textbf{Corresponding Authors}
    \begin{itemize}
        \item Ge Zhang, ByteDance Seed, China and M-A-P
        \item Chenghua Lin, The University of Manchester     
    \end{itemize}

%% file: sections/appendixA.tex
\section{Use of Large Language Models}
\label{appendix:llm_usage}

In accordance with ICLR 2026 policies, we disclose that a large language model was used during the manuscript preparation process to polish and refine the text. The LLM assisted in improving sentence fluency, enhancing clarity of expression, and standardizing language to align with academic writing conventions. All original academic arguments, experimental design, data analysis, and logical structure were developed solely by the authors. The authors independently verified all factual claims and technical content, and take full responsibility for the accuracy and validity of all statements in this paper. 

\section{Full Taxonomy of Writing Categories}
The benchmark spans 51 distinct writing categories, which are grouped into the 8 high-level domains shown below. This comprehensive taxonomy ensures a diverse and representative evaluation of model capabilities across a wide spectrum of writing tasks.

\begin{description}
    \item \textbf{Functional Documents}
        \begin{itemize}
            \item Abstract for Academic Paper 
            \item Experiment Report 
            \item Meeting Minutes 
            \item Resume / Cover Letter 
            \item Thank You / Apology Letter 
            \item Product Manual 
            \item Proposal / Plan 
            \item Interview Questions 
            \item Open Letter 
            \item Argumentative Essay 
            \item Eulogy / Memorial Text 
        \end{itemize}

    \item\textbf{Promotional \& Communication Documents}
        \begin{itemize}
            \item Speech Transcript 
            \item Advertisement Copy / Marketing Email 
            \item Slogan / Tagline 
            \item Social Media Content 
            \item Blog Post 
            \item Product Review 
            \item Popular Science Article 
            \item Tutorial / Guide 
            \item Debate Script 
        \end{itemize}

    \item \textbf{Non-Fiction Writing}
        \begin{itemize}
            \item Prose / Essay 
            \item Biography 
            \item Travelogue 
            \item Book / Film / Music Review 
        \end{itemize}

    \item \textbf{Fiction}
        \begin{itemize}
            \item Fantasy / Magic 
            \item Science Fiction 
            \item Suspense / Mystery 
            \item Historical Story 
            \item Fairy Tale / Fable 
            \item Slice of Life Story 
            \item Emotional / Romance Story 
            \item Wuxia 
            \item Military Fiction 
            \item Historical Fiction (Costume) 
            \item Xuanhuan 
            \item Xianxia 
            \item Gaming Fiction 
            \item Sports Fiction 
            \item General Fiction / Story 
        \end{itemize}

    \item \textbf{Funny}
        \begin{itemize}
            \item ACGN Funny Literature / Doujin (Fan Fiction) 
            \item Fandom Funny Literature 
            \item Esports / Gaming Funny Literature 
            \item Hip-hop / Rap Culture Funny Literature 
            \item Internet Slang Systems 
            \item Anti-Mainstream Consumer Culture 
            \item Cross-national / Cross-lingual Funny Literature 
            \item Subculture Identity Expression 
            \item Role-Playing (as a sub-genre) 
            \item Funny Literature / Subculture 
        \end{itemize}

    \item \textbf{Poetry}
        \begin{itemize}
            \item Poetry 
        \end{itemize}

    \item \textbf{Scriptwriting}
        \begin{itemize}
            \item Play / Script 
        \end{itemize}

    \item \textbf{Role-Playing}
        \begin{itemize}
            \item Role-Playing 
        \end{itemize}
\end{description}

\section{Genre-Specific Scoring Guidelines}
\label{appendix:scoring}

This appendix details the comprehensive framework provided to our expert annotators for evaluating model responses. 
The process is designed to be rigorous and consistent, combining a general quality rubric with a detailed hierarchy of universal and genre-specific standards.

\subsection{General Scoring Rubric (4-Point Scale)}
Each response was assigned a holistic quality score from 0 to 3. The rubric was anchored with descriptive levels and analogies to everyday standards to ensure annotator calibration.

\begin{itemize}
    \item\textbf{Score 3: Creative / Professional}
    The response is creative, stylistically fluent, and feels natural. It is a complete, well-crafted article on par with professionally published work (e.g., in a literary magazine). It is original, engaging, and often exceeds the prompt's expectations in a surprising way.
    
    \item\textbf{Score 2: Competent / Predictable}
    The response is good overall but lacks originality. The structure is sound and the content addresses the prompt, but the narrative or arguments are predictable. This level is analogous to a well-written but standard university-level essay or a competent product manual.

    \item\textbf{Score 1: Formulaic / Flawed}
    The response exhibits significant issues. It may be written in a language different from the one requested, or it follows a rigid, unnatural template (e.g., every paragraph starting with a subheading). The word choice can be awkward or inappropriately technical (e.g., using "quantum" in a non-scientific context). This is comparable to a middle-school-level essay.

    \item\textbf{Score 0: Incoherent / Irrelevant}
    The response is fundamentally unusable. It is nonsensical, completely fails to address the prompt's genre or topic, or consists mostly of a direct repetition of the query. This is analogous to an elementary-school or illiterate level of writing.
\end{itemize}

\subsection{Universal Evaluation Criteria}

Beyond the holistic score, annotators assessed responses against a set of universal criteria applicable to all forms of writing.

\begin{itemize}
    \item \textbf{Prompt Adherence and Intent:}
    \begin{itemize}
        \item Does the response satisfy all explicit constraints in the query (e.g., themes, content, word count)? 
        \item Does it avoid vague, grandiose statements and focus on the core task? 
        \item Is the overall reading experience fluent, not sacrificed for overly ornate or complex sentences? 
    \end{itemize}
    \item \textbf{Structure and Coherence:}
    \begin{itemize}
        \item Is the overall structure complete and are paragraphs divided logically? 
        \item Is the line of reasoning clear and are the ideas logically self-consistent? 
        \item For narratives, is the pacing effective (i.e., a clear beginning, development, climax, and conclusion)? 
    \end{itemize}
    \item \textbf{Content and Substance:}
    \begin{itemize}
        \item Is the content rich and specific, avoiding empty, generic statements?
        \item Does the chosen material effectively support the overall theme or argument?
        \item Where applicable, are environmental descriptions vivid and effective at creating the desired atmosphere?
    \end{itemize}
    \item \textbf{Language and Expression:}
    \begin{itemize}
        \item Is the language accurate, precise, and grammatically correct?
        \item Is the expression clear and unambiguous? 
        \item Does the writing style match the requirements of the prompt, genre, and intended audience? 
    \end{itemize}
\end{itemize}

\subsection{Genre-Specific Evaluation Criteria}

To account for the diverse nature of writing, annotators also applied specific standards for each category. The following are representative examples.

\subsubsection{Fiction (e.g., Sci-Fi, Fantasy, Mystery)}
\begin{itemize}
    \item \textbf{Characters:} Are characters consistent throughout the narrative? Are their relationships (e.g., friendship, rivalry) authentic and do they drive the plot? 
    \item \textbf{Narrative Technique:} Does the author "show" the story through action, dialogue, and detail, rather than simply "telling" the reader what is happening? 
    \item \textbf{Creativity:} Does the story demonstrate originality in its premise, characters, or plot? Does it effectively use narrative devices like foreshadowing and callbacks?
\end{itemize}

\subsubsection{Scriptwriting}
\begin{itemize}
    \item \textbf{Dialogue:} Is the dialogue believable for the characters, reflecting their personality, background, and emotional state? 
    \item \textbf{Action \& Staging:} Does the script include stage directions (e.g., tone, emotion, action) for dialogue? Does it incorporate elements like set design, sound effects, and props? 
    \item \textbf{Motivation:} Do the main characters have clear, understandable motivations that drive the plot forward? 
\end{itemize}

\subsubsection{Non-Fiction (e.g., Essays, Biographies, Reviews)}
\begin{itemize}
    \item \textbf{Accuracy:} Are all factual claims, data, quotes, and historical details accurate? 
    \item \textbf{Authenticity:} Is the author's emotion and experience conveyed in a genuine and credible manner? 
    \item \textbf{Depth:} Does the writing go beyond surface-level description to offer deeper analysis of causes, meanings, or connections? 
\end{itemize}

\subsubsection{Functional Documents (e.g., Resumes, Proposals, Memos)}
\begin{itemize}
    \item \textbf{Purpose:} Is the core purpose of the document (e.g., to inform, persuade, request) immediately clear?
    \item \textbf{Completeness:} Does the document include all information necessary to achieve its goal?
    \item \textbf{Format \& Logic:} Does it follow the conventional format for its type? For persuasive documents, are the arguments clear, well-supported, and logical?
\end{itemize}

\subsubsection{Funny (e.g., Internet Memes, Copypasta)}
This category evaluates a model's grasp of niche, often non-literal communication styles.
\begin{itemize}
    \item \textbf{Form:} Does the response deliberately break conventional logic for humorous or absurd effect?
    \item \textbf{Technique:} Does it correctly use techniques specific to the subculture, such as puns, homophones, context-dependent slang, or "serious nonsense"?
    \item \textbf{Tone:} Does it successfully capture a specific ironic or satirical tone, potentially with multiple layers of meaning?
\end{itemize}

\clearpage

\section{Dataset Statistics}
\label{sec:appendix_statistics}

\begin{table}[h]
\centering
\caption{Distributional properties of preference pairs in \textbf{WritingPreferenceBench}.}
\label{tab:dataset_stats}
\begin{tabular}{@{}llccc@{}}
\toprule
\textbf{Dataset} & \textbf{Metric} & \textbf{Type} & \textbf{Mean (STD)} & \textbf{Median} \\
\midrule
\multirow{4}{*}{English} 
 & \multirow{2}{*}{Length (words)} & Chosen & 1450.3 (1801.9) & 961.5 \\
 & & Rejected & 839.9 (593.4) & 792.0 \\
\cmidrule{2-5}
 & \multirow{2}{*}{Quality Score} & Chosen & 2.913 (0.296) & 3.000 \\
 & & Rejected & 1.602 (0.553) & 2.000 \\
\midrule
\multirow{4}{*}{Chinese} 
 & \multirow{2}{*}{Length (words)} & Chosen & 1873.5 (1967.5) & 1340.5 \\
 & & Rejected & 1458.3 (1311.0) & 1218.5 \\
\cmidrule{2-5}
 & \multirow{2}{*}{Quality Score} & Chosen & 2.560 (0.589) & 3.000 \\
 & & Rejected & 1.115 (0.567) & 1.000 \\
\bottomrule
\end{tabular}
\end{table}




\section{Examples of Benchmark Queries}
To illustrate the nature of the tasks in WritingPreferenceBench, this section provides several examples of the expert-crafted queries given to the models. These queries are designed to be specific, evocative, and challenging, pushing models beyond generic text generation.

\subsection{Example 1: Poetry}
\paragraph{Query:}
\begin{quote}
    Please help me write a modern poem on the theme of the old refrigerator in my grandmother's kitchen. It no longer cools and is now used as a storage cabinet; there are faded stickers and an old shopping list still on its door. The poem needs to start with sensory details like its sound, its smell, and its appearance. It should be depicted as a guardian of family memories. Please use rhetorical devices like personification or metaphor to express a sense of nostalgia and affection for the old days.
\end{quote}

\subsection{Example 2: Product Review}
\paragraph{Query:}
\begin{quote}
    Write a professional product review for a high-end outdoor shell jacket. The article must be well-structured and centered on its core performance metrics. The review should include at least: 1. Design \& Workmanship: Analyze the jacket's fit and cut, fabric technology, seam sealing process, zipper configuration, and overall weight. 2. Core Functionality Test: Objectively evaluate its waterproofing, breathability, and windproofing performance in simulated or real-world conditions. 3. Details \& Usability: Review the hood's range of adjustment, the logic of the pocket layout, and the adjustment systems for the cuffs and hem. The article's conclusion must clearly summarize the product's pros and cons and, in conjunction with its price, provide clear purchasing advice and an analysis of suitable user groups.
\end{quote}

\subsection{Example 3: Funny}
\paragraph{Category:}
\texttt{ACGN Abstract Literature / Doujin (Fan Fiction)}
\paragraph{Query:}
\begin{quote}
    Write an abstract fanfiction piece set against the backdrop of the "Human Instrumentality Project" from Neon Genesis Evangelion. The "protagonist" of this piece is not a specific person, but the very moment in which all human consciousnesses dissolve, merge, and collide within the Sea of LCL. The core theme is "the boundary between the Self and the Other," aiming to explore which is more terrifying: absolute loneliness or the loss of self through fusion. You do not need to construct a plot. Instead, use a fragmented, multi-vocal "stream-of-consciousness" style to weave together the internal monologues, memory fragments, and sensory perceptions of different characters—Shinji's inferiority complex, Asuka's arrogance, Rei's emptiness—and iconic sensory details (like "the metallic taste of orange juice" or "the sweetness of watermelon") to form a chaotic yet harmonious sea of consciousness.
\end{quote}

\subsection{Example 4: Short Story}
\paragraph{Query:}
\begin{quote}
    Write a short story about a conflict between neighbors in a city. The protagonist is a young person who works from home and is constantly bothered by a strange, rhythmic noise coming from their new upstairs neighbor late at night. At the moment they can't stand it anymore and decide to confront the neighbor, they discover an unexpected and poignant truth about the source of the noise. The core of the story is the dramatic turn caused by this revelation, aiming to explore the alienation, misunderstanding, and eventual reconciliation between people in a modern city.
\end{quote}

\subsection{Example 5: Argumentative Essay}
\paragraph{Query:}
\begin{quote}
    On our journey through life, we often face the choice between "looking back" and "moving forward." Some believe that dwelling on the past hinders progress, and thus one must resolutely "move forward." Others are convinced that by frequently "looking back" and drawing wisdom from past experiences and lessons, we can walk the future path more steadily. These two attitudes, seemingly contradictory, are in fact dialectically unified, jointly shaping the trajectory of our lives. Please write an argumentative essay titled "‘Looking Back’ and ‘Moving Forward’". Your viewpoint should be distinct and your essay well-structured. The article must not only discuss why we should "look back" and why we must "move forward," but more importantly, it must delve into how a balance and unity can be achieved between the two. You are required to cite at least one historical figure as positive or negative evidence and analyze it in conjunction with a contemporary social phenomenon or a personal experience. When narrating the case, you must include rich descriptive details that reflect the character's internal journey and emotional changes when faced with the choice between "looking" and "moving." The essay should be no less than 800 words.
\end{quote}

\subsection{Example 6: Speech}
\paragraph{Query:}
\begin{quote}
    You are about to graduate after three years of high school and, as the student representative, you need to deliver a speech at the graduation ceremony. Your audience includes not only the classmates you've spent every day with, but also the hardworking teachers and the parents who have come to attend. Please write a speech manuscript centered on the theme of "Gratitude and Responsibility." The speech should not be a collection of empty slogans or a simple farewell. You must include two specific, detailed stories: first, recount the profound impact a particular teacher had on you, describing a teaching moment or interaction that you still remember vividly; second, share a story of friendship and mutual growth with your classmates. Please create your own title. The speech must be no less than 600 words, with a closing signed by "Graduate Representative, Wang Chen" and the date.
\end{quote}

\subsection{Example 7: Advertising Copy}
\paragraph{Category:}
\texttt{Advertisement Copy / Marketing Email}
\paragraph{Query:}
\begin{quote}
    Write a marketing email for the new "Pathfinder 30L" backpack from the outdoor brand "Nomad's Gear." The email must use a vivid user story (e.g., a summit experience) to highlight the backpack's benefits, such as being lightweight and durable, in order to spark the reader's desire for adventure. A catchy subject line, an immersive story, and a clear call to action at the end are required.
\end{quote}

\subsection{Example 8: Biography}
\paragraph{Query:}
\begin{quote}
    Please help me write a biography of the Mexican painter Frida Kahlo, with a suggested title of "The Burning Thorn Bush: Frida's Pain and Creation." The core theme of this biography should not be a simple chronological account of her life, but an in-depth exploration of "how pain became the core fuel for her artistic creation." You need to closely link and analyze key events in her life (such as the bus accident, her marriage to Diego Rivera, and her miscarriages) with her representative paintings. Provide a detailed interpretation of how she transformed physical disability and emotional turmoil into the powerful symbols and visual language of her artwork.
\end{quote}

%% file: main.bib
@misc{geminiteam2024gemini15unlockingmultimodal,
      title={Gemini 1.5: Unlocking multimodal understanding across millions of tokens of context}, 
      author={Gemini Team},
      year={2024},
      eprint={2403.05530},
      archivePrefix={arXiv},
      primaryClass={cs.CL},
      url={https://arxiv.org/abs/2403.05530}, 
}

@article{pan2022effects,
  title={The effects of reward misspecification: Mapping and mitigating misaligned models},
  author={Pan, Alexander and Bhatia, Kush and Steinhardt, Jacob},
  journal={arXiv preprint arXiv:2201.03544},
  year={2022}
}

@misc{bytedance_doubao15_pro_2025,
  title        = {Doubao-1.5-pro: Exploring the Balance Between Performance and Reasoning},
  author       = {{ByteDance Seed}},
  howpublished = {\url{https://seed.bytedance.com/en/special/doubao_1_5_pro/}},
  year         = {2025},
  note         = {Accessed: 2025-01-22}
}

@article{seed2025seed1,
  title={Seed1. 5-thinking: Advancing superb reasoning models with reinforcement learning},
  author={Seed, ByteDance and Chen, Jiaze and Fan, Tiantian and Liu, Xin and Liu, Lingjun and Lin, Zhiqi and Wang, Mingxuan and Wang, Chengyi and Wei, Xiangpeng and Xu, Wenyuan and others},
  journal={arXiv preprint arXiv:2504.13914},
  year={2025}
}

@misc{bytedance2025seed1.6,
  title        = {Seed 1.6: A General-Purpose Multimodal Model with Adaptive Thinking and a 256K Context Window},
  author       = {ByteDance Seed Team},
  year         = {2025},
  month        = jun,
  howpublished = {\url{https://seed.bytedance.com/en/seed1_6}},
  note         = {Incorporates multimodal capabilities, Adaptive Chain-of-Thought (AdaCoT), supports 23B active parameters / 230B total parameters, 256K context. :contentReference[oaicite:0]{index=0}}
}

@misc{openai2025o3high,
  title        = {OpenAI o3-High (o3 / o3-mini High Reasoning Effort Variant)},
  author       = {{OpenAI}},
  year         = {2025},
  month        = apr,
  note         = {“High reasoning effort” configuration of o3 / o3-mini; see “Introducing OpenAI o3 and o4-mini” and system card},
  howpublished = {\url{https://openai.com/index/introducing-o3-and-o4-mini/}},
  key          = {OpenAI-o3-high}
}

@misc{anthropic2025claude4,
  title        = {Introducing Claude 4: Opus 4 and Sonnet 4},
  author       = {Anthropic},
  year         = {2025},
  howpublished = {\url{https://www.anthropic.com/news/claude-4}},
  note         = {Accessed: 2025-05-23}
}

@article{guo2025deepseek,
  title={Deepseek-r1: Incentivizing reasoning capability in llms via reinforcement learning},
  author={Guo, Daya and Yang, Dejian and Zhang, Haowei and Song, Junxiao and Zhang, Ruoyu and Xu, Runxin and Zhu, Qihao and Ma, Shirong and Wang, Peiyi and Bi, Xiao and others},
  journal={arXiv preprint arXiv:2501.12948},
  year={2025}
}

@misc{qwen2.5,
    title = {Qwen2.5: A Party of Foundation Models},
    url = {https://qwenlm.github.io/blog/qwen2.5/},
    author = {Qwen Team},
    month = {September},
    year = {2024}
}

@misc{fein2025litbenchbenchmarkdatasetreliable,
      title={LitBench: A Benchmark and Dataset for Reliable Evaluation of Creative Writing}, 
      author={Daniel Fein and Sebastian Russo and Violet Xiang and Kabir Jolly and Rafael Rafailov and Nick Haber},
      year={2025},
      eprint={2507.00769},
      archivePrefix={arXiv},
      primaryClass={cs.CL},
      url={https://arxiv.org/abs/2507.00769}, 
}

@article{christiano2017deep,
  title={Deep reinforcement learning from human preferences},
  author={Christiano, Paul F and Leike, Jan and Brown, Tom and Martic, Miljan and Legg, Shane and Amodei, Dario},
  journal={Advances in neural information processing systems},
  volume={30},
  year={2017}
}

@article{zheng2023judging,
  title={Judging llm-as-a-judge with mt-bench and chatbot arena},
  author={Zheng, Lianmin and Chiang, Wei-Lin and Sheng, Ying and Zhuang, Siyuan and Wu, Zhanghao and Zhuang, Yonghao and Lin, Zi and Li, Zhuohan and Li, Dacheng and Xing, Eric and others},
  journal={Advances in neural information processing systems},
  volume={36},
  pages={46595--46623},
  year={2023}
}

@article{lambert2024rewardbench,
  title={Rewardbench: Evaluating reward models for language modeling},
  author={Lambert, Nathan and Pyatkin, Valentina and Morrison, Jacob and Miranda, LJ and Lin, Bill Yuchen and Chandu, Khyathi and Dziri, Nouha and Kumar, Sachin and Zick, Tom and Choi, Yejin and others},
  journal={arXiv preprint arXiv:2403.13787},
  year={2024}
}

@misc{alpaca_eval,
  author = {Xuechen Li and Tianyi Zhang and Yann Dubois and Rohan Taori and Ishaan Gulrajani and Carlos Guestrin and Percy Liang and Tatsunori B. Hashimoto },
  title = {AlpacaEval: An Automatic Evaluator of Instruction-following Models},
  year = {2023},
  month = {5},
  publisher = {GitHub},
  journal = {GitHub repository},
  howpublished = {\url{https://github.com/tatsu-lab/alpaca_eval}}
}

@article{bai2022constitutional,
  title={Constitutional ai: Harmlessness from ai feedback},
  author={Bai, Yuntao and Kadavath, Saurav and Kundu, Sandipan and Askell, Amanda and Kernion, Jackson and Jones, Andy and Chen, Anna and Goldie, Anna and Mirhoseini, Azalia and McKinnon, Cameron and others},
  journal={arXiv preprint arXiv:2212.08073},
  year={2022}
}

@article{ouyang2022training,
  title={Training language models to follow instructions with human feedback},
  author={Ouyang, Long and Wu, Jeffrey and Jiang, Xu and Almeida, Diogo and Wainwright, Carroll and Mishkin, Pamela and Zhang, Chong and Agarwal, Sandhini and Slama, Katarina and Ray, Alex and others},
  journal={Advances in neural information processing systems},
  volume={35},
  pages={27730--27744},
  year={2022}
}

@inproceedings{gao2023scaling,
  title={Scaling laws for reward model overoptimization},
  author={Gao, Leo and Schulman, John and Hilton, Jacob},
  booktitle={International Conference on Machine Learning},
  pages={10835--10866},
  year={2023},
  organization={PMLR}
}

@article{rafailov2023direct,
  title={Direct preference optimization: Your language model is secretly a reward model},
  author={Rafailov, Rafael and Sharma, Archit and Mitchell, Eric and Manning, Christopher D and Ermon, Stefano and Finn, Chelsea},
  journal={Advances in neural information processing systems},
  volume={36},
  pages={53728--53741},
  year={2023}
}

@inproceedings{chiang2024chatbot,
  title={Chatbot arena: An open platform for evaluating llms by human preference},
  author={Chiang, Wei-Lin and Zheng, Lianmin and Sheng, Ying and Angelopoulos, Anastasios Nikolas and Li, Tianle and Li, Dacheng and Zhu, Banghua and Zhang, Hao and Jordan, Michael and Gonzalez, Joseph E and others},
  booktitle={Forty-first International Conference on Machine Learning},
  year={2024}
}

@inproceedings{chodorow2007detection,
  title={Detection of grammatical errors involving prepositions},
  author={Chodorow, Martin and Tetreault, Joel and Han, Na-Rae},
  booktitle={Proceedings of the fourth ACL-SIGSEM workshop on prepositions},
  pages={25--30},
  year={2007}
}

@article{burstein2003finding,
  title={Finding the WRITE stuff: Automatic identification of discourse structure in student essays},
  author={Burstein, Jill and Marcu, Daniel and Knight, Kevin},
  journal={IEEE Intelligent Systems},
  volume={18},
  number={1},
  pages={32--39},
  year={2003},
  publisher={IEEE}
}

@inproceedings{miltsakaki2000automated,
  title={Automated evaluation of coherence in student essays},
  author={Miltsakaki, Eleni and Kukich, Karen},
  booktitle={Proceedings of LREC},
  volume={2000},
  year={2000}
}

@inproceedings{li2018coherence,
  title={Coherence-based automated essay scoring using self-attention},
  author={Li, Xia and Chen, Minping and Nie, Jianyun and Liu, Zhenxing and Feng, Ziheng and Cai, Yingdan},
  booktitle={China National Conference on Chinese Computational Linguistics},
  pages={386--397},
  year={2018},
  organization={Springer}
}

@inproceedings{tay2018skipflow,
  title={Skipflow: Incorporating neural coherence features for end-to-end automatic text scoring},
  author={Tay, Yi and Phan, Minh and Tuan, Luu Anh and Hui, Siu Cheung},
  booktitle={Proceedings of the AAAI conference on artificial intelligence},
  volume={32},
  year={2018}
}

@article{dong2023raft,
  title={Raft: Reward ranked finetuning for generative foundation model alignment},
  author={Dong, Hanze and Xiong, Wei and Goyal, Deepanshu and Pan, Rui and Diao, Shizhe and Zhang, Jipeng and Shum, Kashun and Zhang, Tong},
  journal={arXiv preprint arXiv:2304.06767},
  year={2023}
}

@article{liu2024skywork,
  title={Skywork-reward: Bag of tricks for reward modeling in llms},
  author={Liu, Chris Yuhao and Zeng, Liang and Liu, Jiacai and Yan, Rui and He, Jujie and Wang, Chaojie and Yan, Shuicheng and Liu, Yang and Zhou, Yahui},
  journal={arXiv preprint arXiv:2410.18451},
  year={2024}
}

@article{acemath2024,
  title={AceMath: Advancing Frontier Math Reasoning with Post-Training and Reward Modeling},
  author={Liu, Zihan and Chen, Yang and Shoeybi, Mohammad and Catanzaro, Bryan and Ping, Wei},
  journal={arXiv preprint},
  year={2024}
}

@article{chen2025rm,
  title={Rm-r1: Reward modeling as reasoning},
  author={Chen, Xiusi and Li, Gaotang and Wang, Ziqi and Jin, Bowen and Qian, Cheng and Wang, Yu and Wang, Hongru and Zhang, Yu and Zhang, Denghui and Zhang, Tong and others},
  journal={arXiv preprint arXiv:2505.02387},
  year={2025}
}

@misc{openai2025usage,
  author       = {OpenAI},
  title        = {How People are Using ChatGPT},
  year         = {2025},
  url          = {https://openai.com/index/how-people-are-using-chatgpt/},
  note         = {Accessed: 2025-09-15}
}

@misc{anthropic2025economic,
  author       = {Anthropic},
  title        = {Anthropic Economic Index: Understanding AI’s effects on the economy},
  year         = {2025},
  url          = {https://www.anthropic.com/economic-index},
  note         = {Accessed: 2025-09-16}
}

@article{liu2023alignbench,
  title={Alignbench: Benchmarking chinese alignment of large language models},
  author={Liu, Xiao and Lei, Xuanyu and Wang, Shengyuan and Huang, Yue and Feng, Zhuoer and Wen, Bosi and Cheng, Jiale and Ke, Pei and Xu, Yifan and Tam, Weng Lam and others},
  journal={arXiv preprint arXiv:2311.18743},
  year={2023}
}

@article{wu2025writingbench,
  title={Writingbench: A comprehensive benchmark for generative writing},
  author={Wu, Yuning and Mei, Jiahao and Yan, Ming and Li, Chenliang and Lai, Shaopeng and Ren, Yuran and Wang, Zijia and Zhang, Ji and Wu, Mengyue and Jin, Qin and others},
  journal={arXiv preprint arXiv:2503.05244},
  year={2025}
}
